\documentclass[conference]{IEEEtran} 

\usepackage{graphicx}
\usepackage{mathtools}
\usepackage[percent]{overpic}
\usepackage{amsmath,amsthm,amssymb,upref,color,bm}
\usepackage{calc}
\usepackage{booktabs}
\usepackage{subfig}
\usepackage{hyperref}
\usepackage[noadjust]{cite}

\usepackage{algorithm}
\usepackage{algpseudocode}
\floatname{algorithm}{Algorithm}
\algnewcommand\algorithmicinput{\textbf{Input:}}
\algnewcommand\INPUT{\item[\algorithmicinput]}
\algnewcommand\algorithmicoutput{\textbf{Output:}}
\algnewcommand\OUTPUT{\item[\algorithmicoutput]}
\algrenewcommand\algorithmicrequire{\textbf{Initialize:}}
\algnewcommand\INIT{\item[\algorithmicrequire]}

\DeclareMathOperator*{\argmin}{arg\;min}

\newcommand\vect[1]{\mathbf #1}

\newcommand{\vg}{\vect{g}}

\newcommand{\action}{a}  
  
\newcommand{\timeidx}{t}

\newcommand{\vt}{\vect{t}}
  
\newcommand{\vv}{\vect{v}}  
\newcommand{\vw}{\vect{w}}
\newcommand{\vx}{\vect{x}}  
  
\newcommand{\partition}{\mathcal{F}}

\newcommand{\measlen}{M}

\newcommand{\xsig}{x}

\newcommand{\signalsize}{N}

\newcommand{\edges}{\mathcal{E}}
\newcommand{\cluster}{\mathcal{C}}
\newcommand{\nodes}{\mathcal{V}}
\newcommand{\graph}{\mathcal{G}}

\newcommand{\actionspace}{\mathcal{A}}
\newcommand{\samplingset}{\mathcal{M}}

\newcommand{\actionlist}{\mathcal{L}}
\newcommand\defeq{:=}

\usepackage{setspace}

\title{Graph Signal Sampling via Reinforcement Learning}
\author{\IEEEauthorblockN{Oleksii Abramenko and Alexander Jung}
Department of Computer Science, Aalto University, Finland\\
firstname.lastname@aalto.fi
}

\begin{document}
	\maketitle
\begin{abstract}
We formulate the problem of sampling and recovering clustered graph signal as a multi-armed bandit (MAB) problem. This formulation lends naturally to learning sampling strategies using the well-known gradient MAB algorithm. In particular, the sampling strategy is represented as a probability distribution over the individual arms of the MAB and optimized using gradient ascent. Some illustrative numerical experiments indicate that the sampling strategies based on the gradient MAB algorithm outperform existing sampling methods.\\ 
\end{abstract}

\begin{IEEEkeywords}
 machine learning, reinforcement learning, multi-armed bandit, graph signal processing, total variation, complex networks.
\end{IEEEkeywords}

\section{Introduction}
 \label{sec_intro}
 
Modern information processing systems generate massive datasets which are often strongly heterogeneous, e.g., partially 
labeled mixtures of different media (audio, video, text). A quite successful approach to such 
datasets is based on representing the data as networks or graphs. In particular, we represent datasets by graph signals
defined over an underlying graph, which reflects similarities between individual data points.
The graph signal values encode label information which often conforms to a clustering hypothesis, i.e., the signal values (labels) of close-by nodes (similar data points) are similar. 

Two core problems considered within graph signal processing (GSP) are (i) how to sample them, i.e., 
which signal values provide the most information about the entire dataset, and (ii) how to 
recover the entire graph signal from these few signal values (samples). 
These problems have been studied in \cite{EslamlouJung, jung2016scalable, jung2017network, JungTranMara2018, sharpnack2012sparsistency, MaraAsilomar2017} which discussed convex 
optimization methods for recovering a graph signal from a small number of signal values observed on the nodes belonging to a given (small) sampling set. Sufficient conditions on the sampling set and clustering structure such that these convex methods are successful have been discussed in \cite{JungTranMara2018, AlexMadelon2017}.
 
{\bf Contribution.} We propose a novel approach to the graph signal sampling and recovery it by interpreting it as a reinforcement learning (RL) problem. In particular, we interpret online sampling algorithm as an artificial intelligence agent which chooses the nodes to be sampled on-the-fly. The behavior of the sampling agent is represented by a probability distribution (``policy'') over a discrete set of different actions which are at the disposal of the sampling agent in order to choose the next node at which the graph signal is sampled. The ultimate goal is to learn a sampling policy which chooses signal samples that allow for a small reconstruction error.

{\bf Notation.} 
The vector with all entries equal to zero is denoted $\mathbf{0}$. Given a vector $\vx$ with non-negative entries, we denote by $\sqrt{\mathbf{x}}$ the vector whose entries are the square roots of the entries of $\vx$. Similarly, we denote the element-wise square of the vector as $\vx^{2}$.

{\bf Outline.} In Section \ref{sec_setup} we formulate the problem of recovering a clustered graph signal from its values on few nodes forming a sampling set as a convex optimization problem. 
Our main contribution is in Section \ref{sec_contribution} where 
we introduce our RL-based sampling method. The results of some numerical experiments 
are presented in Section \ref{sec4_numerical}. We discuss our findings in Section \ref{sec_discussion} and finally conclude in Section \ref{sec_conclusion}. 

\section{Problem Formulation}
\label{sec_setup}

We consider datasets which are represented by a data graph $\graph=(\nodes,\edges)$. 
The data graph is an undirected connected graph (no self-loops and no multi-edges) with nodes $\nodes=\{1,\ldots,\signalsize\}$, which are connected by edges $\{i,j\} \in \edges$. Each node $i\in \nodes$ represents an individual data point and an edge $\{i,j\} \in \edges$ 
connects nodes representing similar data points.
The distance ${\rm dist}(i,j)$ between two different nodes $i,j \in \nodes$ is defined as the length of the shortest path 
between those nodes. 
For a given node $i \in \nodes$, we define its neighbourhood as 
\begin{equation*}
\mathcal{N}(i) \defeq \{ j \in \nodes: \{i,j\} \in \edges \}. 
\end{equation*} 
It will be handy to generalize the notion of neighbourhood and define, for some $r  \in \mathbb{N}$, the $r$-step 
neighbourhood of a node $i \in \nodes$ as $\mathcal{N}(i,r) \defeq \{ j \in \nodes: {\rm dist}(i,j) = r \}$. The 1-step neighbourhood coincides with the neighbourhood of a node, i.e., $\mathcal{N}(i,1) = \mathcal{N}(i)$.  

In many applications we can associate each data point $i\in \nodes$ with a label $x[i] \in \mathbb{R}$. 
These labels induce a graph signal $\vx: \nodes \rightarrow \mathbb{R}$ defined over the graph $\graph$.

We aim at recovering a graph signal $\vx$ based on observing its values $x[i]$ only for nodes $i$ belonging to a sampling set  
\begin{equation*} 
\samplingset \defeq\{i_{1},\ldots,i_{\measlen}\} \subseteq \mathcal{V}.
\end{equation*}
Since acquiring signal values (i.e., labelling data points) is often expensive (requiring manual labor), the sampling set is typically much smaller than the overall dataset, i.e., $\measlen=|\samplingset| \ll \signalsize$. For 
a fixed sampling set size (sampling budget)  $\measlen$ it is important to choose the sampling 
set such that the signal samples $\{ x[i] \}_{i \in \samplingset}$ carry maximal information about the 
entire graph signal. 

The recovery of the entire graph signal from (few) signal samples $\{ x[i] \}_{i \in \samplingset}$  is possible for clustered graph signals 
which do not vary too much over well-connected subset of nodes (clusters) (cf. \cite{JungTranMara2018,JungHero2016}). 
We will quantify how well a graph signal is aligned to the cluster structure using the total variation (TV) 
\begin{equation*} 
\label{equ_def_TV}
\| \vx \|_{\rm TV} \defeq \sum_{\{i,j\} \in \edges} |x[j]\!-\!x[i]|. 
\end{equation*} 
Recovering a graph signal based on the signal values $x[i]$ for the nodes $i \in \samplingset$ of the sampling set $\samplingset$ can be accomplished by solving
\begin{align}
\label{equ_optim_prob}
\hat{\vx}^{\mathcal{M}} &\!\in\! \argmin\limits_{\tilde{\vx}} \| \tilde{\vx} \|_{\rm TV} \,\, {\rm s.t. } \,\,  \tilde{x}[i]\!=\!x[i] \mbox{ for all } i\!\in\!\samplingset.
\end{align}
This is a convex optimization problem with a non-differentiable objective function which precludes the use 
of simple gradient descent methods. However, by applying the primal-dual method of Pock and Chambolle \cite{PockChambolle} 
to solve the recovery problem \eqref{equ_optim_prob}, an efficient sparse label propagation algorithm has been obtained in \cite{JungHero2016}.

A simple but useful model for clustered graph signals is:
\begin{equation}
\label{equ_clust_gsig}
\vx = \sum_{\cluster \in \partition} a_{\cluster} \vt_{\cluster}, 
\end{equation}
with the cluster indicator signals
\begin{equation*} 
\label{equ_def_indicator_signal}
t_{\cluster}[i] = \begin{cases} 1 \mbox{, if }  i \in \cluster \\ 0 \mbox{ else.}  \end{cases}
\end{equation*}
The partition $\partition$ underlying the signal model (\ref{equ_clust_gsig}) can be chosen arbitrarily in principle. However, our methods are expected to be most useful if the partition matches the intrinsic cluster structure of the data graph $\graph$. The clustered graph signals of the form \eqref{equ_clust_gsig} conform with the network topology, in the sense of having small TV $\|�\vx \|_{\rm TV}$, 
if the underlying partition $\partition=\{\cluster_{1},\ldots,\cluster_{|\partition|}\}$ consists of disjoint clusters $\cluster_{l}$ with small cut-sizes. 
Relying on the clustered signal model \eqref{equ_clust_gsig}, \cite[Thm. 3]{AlexMadelon2017} presents a sufficient condition on the 
choice of sampling set such that the solution $\hat{\vx}$ of \eqref{equ_optim_prob} coincides with the true underlying clustered graph signal of the form \eqref{equ_clust_gsig}. 
The condition presented in \cite[Thm. 3]{AlexMadelon2017} suggests to choose the nodes in the sampling set preferably near the boundaries between the different clusters.

\section{Signal Sampling as Reinforcement Learning}
\label{sec_contribution}

The problem of selecting the sampling set $\samplingset$ and recovering the entire graph signal $\vx$ from the signal values $x[i]$ can be interpreted as a RL problem. Indeed, we consider the selection of the nodes to be sampled being carried out by an ``agent'' which 
crawls over the data graph $\graph$. The set of actions our sampling agent may take is $\actionspace = \{1,\ldots,H\}$. 

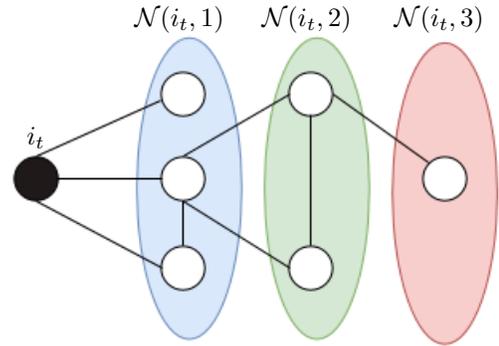
\begin{figure}
	\centering
	\hspace*{0.5em} \begin{overpic}[scale = 0.8]{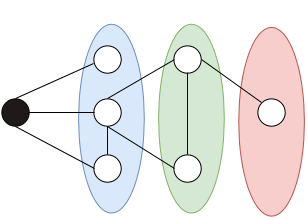}
	\put (3,42) {$i_{t}$}
	\put (25,66) {$\mathcal{N}(i_{t}, 1)$}
	\put (51,66) {$\mathcal{N}(i_{t}, 2)$}
	\put (78,66) {$\mathcal{N}(i_{t}, 3)$}
	\end{overpic}
	\caption{The filled node represents the current location $i_{t}$ of the sampling agent at time $t$. We also indicate the $1$-, $2$- and $3$-step neighbourhoods.}
\label{fig:Neighbourhoods}
\vspace{-5mm}
\end{figure}

A specific action $a \in \actionspace$ refers to the number of hops the sampling agent performs starting at the current node $i_{t}$  to reach a new node $i_{t+1}$, which will be added to the sampling set, i.e., $\samplingset \defeq \samplingset \cup \{ i_{t+1} \}$. In particular, the new node $i_{t+1}$ is selected uniformly at random among the nodes which belong to its $a$-step neighbourhood $\mathcal{N}(i_{t},a)$ (see Figure \ref{fig:Neighbourhoods}).

The problem of optimally selecting actions at given time can be formulated as a MAB problem. Each arm of the bandit is associated with an action. In our setup, a sampling strategy (or policy) amounts to specifying a probability distribution over the individual actions $\action \in \actionspace$. We parametrize this probability distribution with a weight vector $\vw = (w_{1},\ldots,w_{H}) \in \mathbb{R}^{H}$ using the softmax rule:
\begin{equation*}
\pi^{(\vw)}(\action) = \frac{e^{w_{\action}}}{\sum\limits_{b \in \actionspace}{e^{w_{b}}}}
\label{softmax_prob}
\end{equation*}

The weight vector $\vw$ is tuned in the episodic manner with each episode amounting to selecting sampling set $\samplingset$ based on the policy $\pi^{(\vw)}$. At each timestep $\timeidx$ the agent randomly draws an action $a_\timeidx$ according to the distribution $\pi^{(\vw)}$ and performs transition to the next node $i_{\timeidx+1}$ which is selected uniformly at random from the $a_\timeidx$-step neighbourhood $\mathcal{N}(i_{t},a_\timeidx)$. As was mentioned earlier, the node $t_{\timeidx+1}$ is added to the sampling set, i.e., $\samplingset \defeq \samplingset \cup \{ i_{t+1} \}$. We also record the action $a_{\timeidx}$ and add it to the action list, i.e., $\actionlist \defeq \actionlist \cup \{ \action_t \}$. The process continues until we obtain a sampling set $\samplingset$ with the prescribed size (sampling budget) $M$.

Our goal is to learn an optimal policy  $\pi^{(\vw)}$ for the sampling agent in order to obtain signal samples which allow recovery of the entire graph signal with minimum error. We assess the quality of the policy using the mean squared error (MSE) incurred by the recovered signal $\hat{\vx}^{\samplingset}$ which is obtained via (\ref{equ_optim_prob}) using the sampling set $\samplingset$ by following policy $\pi^{(\vw)}$:

\begin{equation*}
R:= -\frac{1}{\signalsize}\sum\limits_{j \in \nodes}(\xsig[j] - \hat{\xsig}^{\samplingset}[j])^2
\end{equation*}

The obtained reward is associated with all actions/arms which contributed to picking samples into sampling set during the episode. For example, if the sampling set has been obtained by pulling arms 1, 2 and 5, the obtained reward will be associated with all these arms, because we do not know what is the exact contribution of the specific arm to the finally obtained MSE. 

The key idea behind gradient MAB is to update weights $\vw$ so that actions yielding higher rewards become more probable under $\pi^{(\vw)}$ \cite[Chapter 2.8]{ref:SuttonBarto}. According to the aforementioned book weights update can be accomplished using gradient ascent algorithm:
\begin{equation}
\begin{aligned}
&w_{\action} \defeq  \begin{cases}
w_{\action} + \alpha R(1-\pi(a)),\action = a_k\\
w_{\action} - \alpha R\pi(\action),\forall \action \neq a_k
\end{cases}\\
&\text{for } k = 1..M-1, \action \in \actionspace, \action_k \in \actionlist
\end{aligned}
\label{update_rule}
\end{equation}

The single difference between update rule (\ref{update_rule}) and one presented in the book \cite[Eq. 2.10]{ref:SuttonBarto} is that in our case weights update is performed in the end of each episode and not after an arm pull. That is because we do not know reward immediately after pulling an arm and should wait until the whole sampling set is collected and reward is observed. 
The intuition behind the update equation (\ref{update_rule}) is that for each arm which has participated in picking a node into sampling set ($\action = a_t$), the weight is increased, whereas weights of remaining arms ($\forall \action \neq a_t$) are decreased. In both cases degree of weight increase/decrease is scaled by the reward obtained with help of this arm as well as by the learning rate $\alpha$. For faster convergence in our implementation, instead of stochastic gradient ascent we use mini-batch gradient ascent in combination with RMSprop technique  \cite{hiltonRMSprop} (see Algorithm \ref{algo_bandit} for implementation details).

\begin{figure}%
    \centering
    \subfloat[]{{\includegraphics[width=3.3cm]{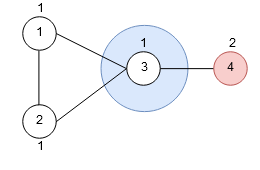} }}%
    \qquad
    \subfloat[]{{\includegraphics[width=3.3cm]{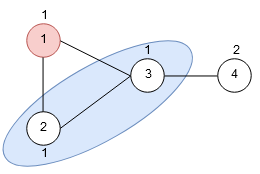} }}%
    \caption{Illustration of reward being conditioned on the position of a sampling agent. In this picture: red node -- current position of the sampling agent, blue region -- nodes within distance 1 from the sampling agent. Node indices are shown inside the nodes, signal values -- outside.}
    \label{fig:reward_non_stat}%
\vspace{-5mm}
\end{figure}

Choice of the gradient MAB algorithm can be additionally justified by the study \cite{besbes2014} which shows that in the environments with non-stationary rewards probabilistic MAB policy can result in higher expected reward in comparison to single-best action policies. In our problem non-stationarity of reward arises from the graph structure itself, i.e., reward distribution for a particular arm of a bandit depends on the location of the sampling agent. Suppose sampling budget $M$ is 2 and consider example presented in Figure \ref{fig:reward_non_stat}. In case (a) sampling agent is initially located at node 4. By pulling arm \#1 it can only pick node 3 which is in the other cluster. It is easy to verify that by using recovery method (\ref{equ_optim_prob}) graph signal will be perfectly reconstructed (MSE = 0). On the other hand, case (b) shows the situation when the agent can only pick nodes 2 or 3 belonging to the same cluster as currently sampled node, leading to non-zero reconstruction MSE.

The whole process of weight updates is repeated for sufficient number of episodes until convergence is reached and the optimal stochastic policy is attained. Described above learning procedure can be efficiently summarized in the form of pseudocode (see Algorithm \ref{algo_bandit}).

\begin{algorithm}
    \caption{Online Sampling and Reconstruction}
    \begin{algorithmic}[1]
        \Statex 
        \INPUT {data graph $\graph$, sampling budget $\measlen$, batch size $B$, $\alpha$}
         \INIT { $\vw : = \mathbf{0}, \samplingset = \{\emptyset\}, \actionlist = \{\emptyset\}, \nabla \vw = \mathbf{0}, \vg = \mathbf{0},  ep = 0$}
            \Repeat
                \State select starting node $i \in \nodes$ randomly 
                \State $\samplingset := \{i \}$
                \State $\actionlist = \{\emptyset\}$
                \For{\texttt{$\timeidx := 1 ; \timeidx < \measlen$}}
                    \State  $\action:= \Call{SampleAction}{\pi^{(\vw)}}$
                       \State  $i_{next} :=\Call{SampleNode}{\graph, \mathcal{N}(i,a)}$
                       \State $\samplingset := \samplingset \cup \{i_{next} \}$
                       \State $\actionlist := \Call{AppendToList}{\actionlist, \action}$
                         \State  $i := i_{next}$
                \EndFor
                
                \State $\hat{\vx} \!\in\! \argmin\limits_{\tilde{\vx}} \| \tilde{\vx} \|_{\rm TV}$
                \State${\rm s.t. } \,\,  \tilde{x}[i]\!=\!x[i] \mbox{ for all } i\!\in\!\samplingset$
                \State $R:= -\frac{1}{\signalsize}\sum\limits_{j \in \nodes} (\xsig[j] - \hat{\xsig}[j])^2$
                \For{\texttt{$k := 1 ; k < \measlen$}}
                    \For{\texttt{$a := 1 ; a \leqslant H$}}
                       \State $\begin{aligned}
                        &\nabla w_{\action} \defeq  \begin{cases}
                        \nabla w_{\action} + R(1-\pi(a))\text{, if } \action = \actionlist[k]\\
                        \nabla w_{\action} - R\pi(\action)\text{, otherwise}
                        \end{cases}\\
                        \end{aligned}$
                    \EndFor
                \EndFor
                
                \State $ep \defeq ep + 1$
                
                 \If{$ep$ mod $B = 0$}
                 \State $\vg \defeq 0.9 \vg + 0.1 (\nabla \vw)^2 $
                 \State $\vw \defeq \vw +  \alpha\nabla \vw/\sqrt{\vg}$
                 \State $\nabla \vw \defeq \mathbf{0}$
                 \EndIf
                
            \Until convergence is reached
        \OUTPUT {$\pi^{(\vw)}$} 
    \end{algorithmic}  
    \label{algo_bandit}
\end{algorithm}

Obtained probability distribution $\pi^{(\vw)}$ represents sampling strategy which incurs the minimum reconstruction MSE when using the convex recovery method (\ref{equ_optim_prob}).

\section{Numerical Results} 
\label{sec4_numerical}

\begin{figure}
	\centering
	\includegraphics[width = \columnwidth, height=6cm]{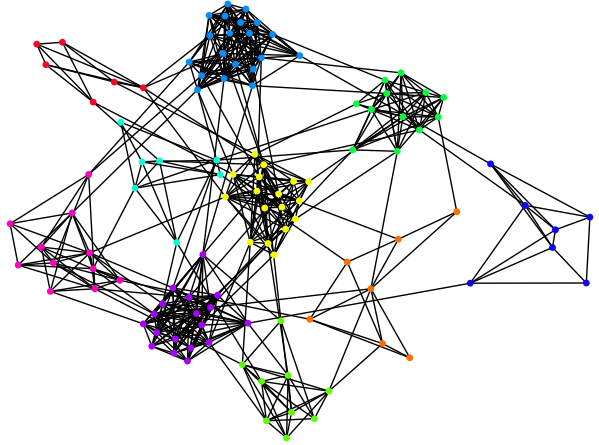}
	\caption{Data graph obtained from the stochastic block model with $p=7/10$ and $q=1/100$.}
	\label{fig:Graph1}
\vspace{-5mm}
\end{figure}

We now verify the effectiveness of the proposed sampling set selection algorithm using synthetic data and compare it to two other existing approaches, i.e., random  walk
sampling (RWS) \cite{ref:RandomWalksampling} and uniform random sampling (URS) \cite[Section 2.3]{puy2016random}. We define a random graph with 10 clusters where sizes of clusters are drawn from the geometric distribution with probability of success $8/100$. In accordance to the stochastic block model (SBM) \cite{ref:mossel2012} intra- and inter-cluster connection probabilities are parametrized as $p = 7/10$ and $q = 1/100$. We then generate a clustered graph signal according to (\ref{equ_clust_gsig}) with the signal coefficients $a_{\cluster_{l}}= l$ for $l = 1,2,...,10$. Example of a typical instance of random graph with such parameters is shown in Figure \ref{fig:Graph1}.

\begin{figure}
	\centering
	\includegraphics[scale = 0.57]{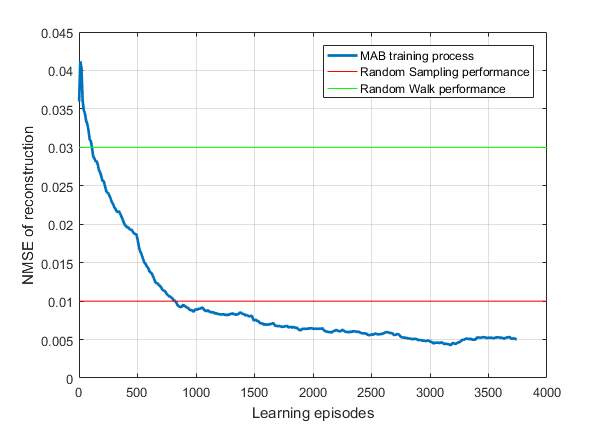}
	\caption{Convergence of gradient MAB for one learning instance $\graph_i$ (showing first 3700 episodes).}
	\label{fig:conv}
\vspace{-5mm}
\end{figure}

Given the model we generate training data consisting of $K = 500$ random graphs and for each graph instance we run Algorithm $\ref{algo_bandit}$ for 10000 episodes, which is sufficient to reach convergence. It is interesting to note that the algorithm outperforms RWS and URS strategies after 200 and 800 episodes respectively (see Figure \ref{fig:conv}). Convergence speed is high at the initial stage and then substantially decreases after approximately 1000 episodes.  

\begin{figure}
	\centering
	\includegraphics[width = \columnwidth, height=6cm]{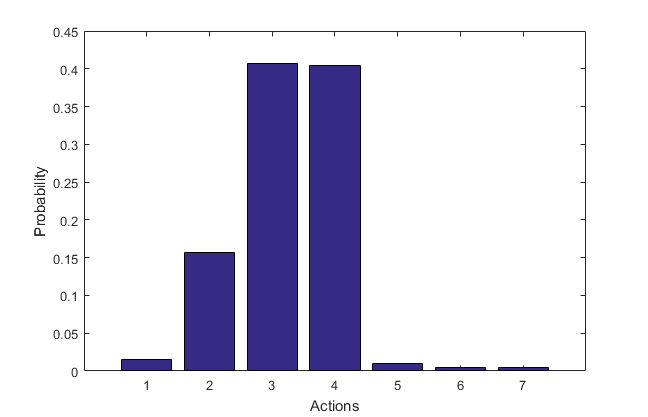}
	\caption{Mean policy for the stochastic block model family $\graph$.}
	\label{fig:policy}
\vspace{-5mm}
\end{figure}

In Figure \ref{fig:policy} we illustrate the mean policy

\begin{equation}
\pi^{(\vw)} = \frac{1}{K}\sum_{i=1}^{K} \pi^{(\vw)}_{i}
\label{ref:average_distribution}
\end{equation}

The finally obtained policy (\ref{ref:average_distribution}) is then evaluated by applying it to 500 new i.i.d. realizations of the data graph, yielding the sampling sets $\samplingset^{(i)}$, $i=1,\ldots,500$, and measuring the normalized mean squared error (NMSE) incurred by graph signal recovery from those sampling sets:

\begin{equation*}
NMSE_{\graph_i} = \frac{\|\hat{\xsig}^{(i)} - \xsig^{(i)}\|_{2}^{2}}{\|\xsig^{(i)}\|_{2}^{2}}
\end{equation*}

\begin{equation*}
NMSE = \frac{1}{500}\sum_{i=1}^{500} NMSE_{\graph_i}
\end{equation*}

We perform similar measurements of the NMSE for random walk and random sampling algorithms under different sampling budgets and convert results to the logarithmic scale.



\begin{figure}
	\centering
	\includegraphics[scale = 0.51]{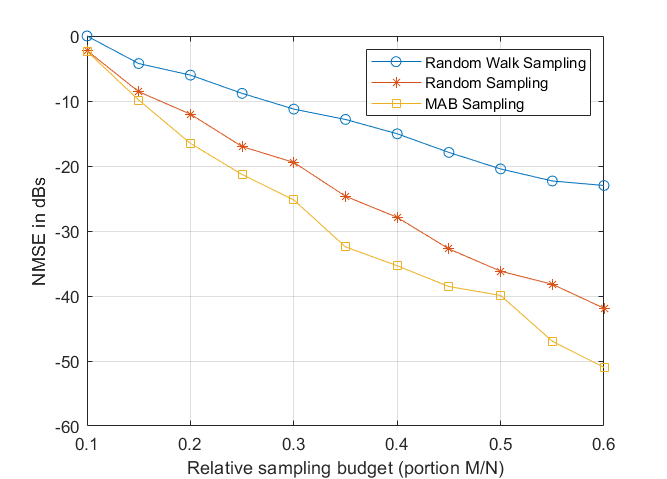}
	\caption{Test set error obtained from graph signal recovery based on different sampling strategies.}
	\label{fig:Results_1}
\vspace{-5mm}
\end{figure}

The Figure \ref{fig:Results_1} shows that for relative sampling budget 0.2 improvement in NMSE amounts to 5 dBs in comparison to random sampling and 10 dBs in comparison to random walk approach. This gap increases even more for the sampling budget 0.4, to 8 dBs and 20 dBs respectively. The general tendency suggests further increase of the gap for larger sampling budgets.

\section{Discussion}
\label{sec_discussion}
We now interpret the results and explain the poor performance of RWS using a simple argument based on the properties of Markov chains. For simplicity we consider a graph with clusters $\cluster_{1}$ and $\cluster_{2}$ having sizes $N_1$ and $N_2$. The probability of having an edge between nodes in the same cluster is denoted $p$, while the probability of having an edge between nodes in different clusters is $q$. An elementary calculation yields the probability of a random walk transitioning from $\cluster_{1}$ to $\cluster_{2}$ as:

\begin{equation*}
p_{12} = \frac{qN_2}{qN_2 + p(N_1 - 1)}
\end{equation*}

Likewise, the probability of staying in the $\cluster_{1}$:
\begin{equation*}
p_{11} = 1 - p_{12}
\end{equation*}

We note that $qN_2$ is the expected number of edges between a particular node of $\cluster_{1}$ and $\cluster_{2}$ and $p(N_1 - 1)$ is the expected number of edges between a particular node of $\cluster_{1}$ and the remaining nodes of $\cluster_{1}$. Similarly for $\cluster_{2}$:

\begin{equation*}
p_{21} = \frac{qN_1}{qN_1 + p(N_2 - 1)}
\end{equation*}
\begin{equation*}
p_{22} = 1 - p_{21}
\end{equation*}

The transition matrix of a Markov chain, which summarizes probabilistic transitions between clusters, can be formalized as follows:

\begin{equation*}
    P = 
    \begin{bmatrix}
        p_{11}       & p_{12}\\
        p_{21}       & p_{22}
    \end{bmatrix}
\end{equation*}
Let $\vv = (v_1, v_2)^T$ be an equilibrium distribution \cite{norris1998markov} of the Markov chain which reflects amount of discrete time spent in $\cluster_{1}$ and $\cluster_{2}$. According to theory of Markov chains \cite{norris1998markov} finding this distribution amounts to finding a vector $\vv$ such that:
\begin{equation}
    \begin{cases}
    \vv^{T}P = \vv^{T}\\
    v_1 + v_2 = 1\\
    v_1 \geqslant 0, v_2 \geqslant 0\\
    \end{cases} 
\label{system_eq}
\end{equation}

It is easy to verify that solving the aforementioned system (\ref{system_eq}) yields the following equilibrium distribution:
\begin{equation*}
v_1 = \frac{p_{21}}{p_{12} + p_{21}}, v_2 = 1-v_1
\end{equation*}

We now consider particular example of a random graph with the configuration: $N_1 = 20,N_2 = 80,p = 0.7, q = 0.01$. According to the presented above formulas, computations yield equilibrium distribution: $v_1 \approx 0.05$, $v_2 \approx 0.95$, which means that 95 \% of discrete time of a random walk is spent in $\cluster_{2}$ whereas only 5\% of time is spent in $\cluster_{1}$. This rationale implies that upon termination of a random walk instance its endpoints will be located in clusters $\cluster_{1}$ and $\cluster_{2}$ with probabilities $0.05$ and $0.95$ respectively.

From the aforementioned examples we can conclude that although $\cluster_{2}$ is only four times larger than $\cluster_{1}$, the probability of random walk termination within it is larger by a factor $\approx 19$. Thus, the random walk sampling algorithm tends to oversample larger clusters and undersample smaller ones. This partially explains the poor performance of random walk in comparison to random sampling which samples clusters proportionally to their sizes.

\section{Conclusions}
\label{sec_conclusion}
This paper proposes a novel approach for graph signal processing which is based on interpreting graph signal sampling and recovery as a reinforcement learning problem. Using the lens of reinforcement learning lends naturally to an online sampling strategy which is based on determining an optimal policy which minimizes MSE of graph signal recovery. The proposed approach has been tested on a synthetic dataset generated in accordance to the stochastic block model. Obtained experimental results have confirmed effectiveness of the proposed sampling algorithm in the stochastic settings and demonstrated its advantages over existing approaches.

\vspace*{1mm}
\bibliographystyle{IEEEtran}
\bibliography{IEEEabrv,references}

\end{document}